\documentclass[10pt,twocolumn,letterpaper]{article}

\usepackage{iccv}

\usepackage{times}
\usepackage{epsfig}
\usepackage{graphicx}
\usepackage{amsmath}
\usepackage{amssymb}

\usepackage{xcolor}
\usepackage{algorithm}
\usepackage{listings}
\usepackage[british,american]{babel}
\usepackage{enumitem}
\usepackage{multirow}

\usepackage{textpos}  

\definecolor{mygray}{gray}{0.6}
\newcommand{\demph}[1]{\textcolor{mygray}{#1}}

\newcommand{\app}{\raise.17ex\hbox{$\scriptstyle\sim$}}

\usepackage{tabulary}
\newcolumntype{x}[1]{>{\centering\arraybackslash}p{#1pt}}

\usepackage{etoolbox}
\makeatletter
\AfterEndEnvironment{algorithm}{\let\@algcomment\relax}
\AtEndEnvironment{algorithm}{\kern2pt\hrule\relax\vskip3pt\@algcomment}
\let\@algcomment\relax
\newcommand\algcomment[1]{\def\@algcomment{\footnotesize#1}}
\renewcommand\fs@ruled{\def\@fs@cfont{\bfseries}\let\@fs@capt\floatc@ruled
  \def\@fs@pre{\hrule height.8pt depth0pt \kern2pt}%
  \def\@fs@post{}%
  \def\@fs@mid{\kern2pt\hrule\kern2pt}%
  \let\@fs@iftopcapt\iftrue}
\makeatother

\definecolor{citecolor}{HTML}{0071bc}
\usepackage[pagebackref=true,breaklinks=true,letterpaper=true,colorlinks,citecolor=citecolor,bookmarks=false]{hyperref}

\newcommand\blfootnote[1]{%
  \begingroup
  \renewcommand\thefootnote{}\footnote{#1}%
  \addtocounter{footnote}{-1}%
  \endgroup
}
\newlength\savewidth\newcommand\shline{\noalign{\global\savewidth\arrayrulewidth
  \global\arrayrulewidth 1pt}\hline\noalign{\global\arrayrulewidth\savewidth}}
\newcommand{\tablestyle}[2]{\setlength{\tabcolsep}{#1}\renewcommand{\arraystretch}{#2}\centering\footnotesize}
\makeatletter\renewcommand\paragraph{\@startsection{paragraph}{4}{\z@}
  {.5em \@plus1ex \@minus.2ex}{-.5em}{\normalfont\normalsize\bfseries}}\makeatother

\iccvfinalcopy 

\def\iccvPaperID{****} 

\begin{document}

\title{\vspace{-2em} An Empirical Study of Training Self-Supervised Vision Transformers \vspace{-.5em}}

\author{Xinlei Chen$^{*}$\qquad Saining Xie$^{*}$\qquad Kaiming He\\
Facebook AI Research (FAIR)
\vspace{0em}
}

\maketitle
\ificcvfinal\thispagestyle{empty}
\blfootnote{*: equal contribution.}\fi

\vspace{-1.5em}
\begin{abstract}
\vspace{-.5em}
This paper does not describe a novel method. Instead, it studies a straightforward, incremental, yet must-know baseline given the recent progress in computer vision: self-supervised learning for Vision Transformers (ViT).
While the training recipes for standard convolutional networks have been highly mature and robust, the recipes for ViT are yet to be built, especially in the self-supervised scenarios where training becomes more challenging.
In this work, we go back to basics and investigate the effects of several fundamental components for training self-supervised ViT. We observe that instability is a major issue that degrades accuracy, and it can be hidden by apparently good results.
We reveal that these results are indeed partial failure, and they can be improved when training is made more stable.
We benchmark ViT results in \mbox{MoCo v3} and several other self-supervised frameworks, with ablations in various aspects. We discuss the currently positive evidence as well as challenges and open questions. We hope that this work will provide useful data points and experience for future research.
\vspace{-2em}
\end{abstract}

\begin{textblock*}{.8\textwidth}[.5,0](0.5\textwidth, -.535\textwidth)
\centering
{\small Code: \url{https://github.com/facebookresearch/moco-v3}}
\end{textblock*}

\section{Introduction} \label{sec:intro}

Unsupervised pre-training has revolutionized natural language processing (NLP) \cite{Radford2018,Devlin2019,Radford2019,Brown2020}. In computer vision, the un-/self-supervised pre-training paradigms differ from their NLP counterparts in at least two aspects:
(i) the learners in NLP are masked auto-encoders, while in vision the recently popular choices are Siamese networks (\eg, \cite{He2020,Chen2020,Grill2020,Caron2020}); 
(ii) the backbone architectures in NLP are self-attentional Transformers \cite{Vaswani2017}, while in vision the common choice is convolutional \cite{LeCun1989}---yet non-attentional---deep residual networks (ResNets) \cite{He2016}. To complete the big picture of self-supervised learning in vision, and towards closing the gap of pre-training methodology between vision and language, it is of scientific merit to investigate these differences.

This work focuses on training Transformers with the leading self-supervised frameworks in vision.
This investigation is a straightforward extension given the recent progress on Vision Transformers (ViT) \cite{Dosovitskiy2021}. In contrast to prior works \cite{Chen2020c,Dosovitskiy2021} that train self-supervised Transformers with masked auto-encoding, we study the frameworks that are based on Siamese networks, including MoCo \cite{He2020} and others \cite{Chen2020,Grill2020,Caron2020}.

Unlike standard convolutional networks whose training practice has been extensively studied thanks to continuous community effort, ViT models are new and their recipes are yet to be established. In this work, we go back to basics and investigate the fundamental components of training deep neural networks: the batch size, learning rate, and optimizer. We find that under various cases, instability is a major issue that impacts self-supervised ViT training.

Interestingly, we observe that unstable ViT training may \emph{not} result in catastrophic failure (\eg, divergence); instead, it can cause \emph{mild} degradation in accuracy (\eg, 1$\app$3\%). Such a degree of degradation may not be noticeable, unless a more stable counterpart is available for comparison. To the best of our knowledge, this phenomena is rare in the literature of training convolutional networks\footnotemark, and we believe this problem and its hidden degradation are worth noticing.

\footnotetext{See also postscript on a related discussion.}

\begin{table}[t]
\centering
\vspace{-.3em}
\tablestyle{8pt}{1.1}
\begin{tabular}{llrc}
framework & model & params & acc. (\%) \\
\shline
\textit{linear probing:} \\ 
iGPT \cite{Chen2020c} & iGPT-L & 1362M & 69.0 \\
iGPT \cite{Chen2020c} & iGPT-XL & 6801M & 72.0 \\
MoCo v3 & ViT-B & 86M & 76.7 \\
MoCo v3 & ViT-L & 304M & 77.6 \\
MoCo v3 & ViT-H & 632M & 78.1 \\
MoCo v3 & ViT-BN-H & 632M & 79.1 \\
MoCo v3 & ViT-BN-L/7 & 304M & \textbf{81.0} \\
\hline
\textit{end-to-end fine-tuning:} \\ 
{masked patch pred. \cite{Dosovitskiy2021}} & {ViT-B} & {86M} & {79.9}$^\dagger$ \\
MoCo v3 & ViT-B & 86M & {83.2} \\
MoCo v3 & ViT-L & 304M & \textbf{84.1} \\
\end{tabular}
\vspace{.5em}
\caption{\hspace{1em}\textbf{State-of-the-art Self-supervised Transformers} in \mbox{ImageNet} classification, evaluated by linear probing (top panel) or end-to-end fine-tuning (bottom panel).
Both iGPT \cite{Chen2020c} and masked patch prediction \cite{Dosovitskiy2021} belong to the masked auto-encoding paradigm. \mbox{MoCo~v3} is a contrastive learning method that compares two (224$\times$224) crops.
\mbox{ViT-B}, -L, -H are the Vision Transformers proposed in \cite{Dosovitskiy2021}.
\mbox{ViT-BN} is modified with BatchNorm, and ``/7'' denotes a patch size of 7$\times$7. $^\dagger$: pre-trained in JFT-300M.
\label{tab:sota_ssl_transformers}
}
\vspace{-1em}
\end{table}

To demonstrate the possible harm of instability, we investigate a simple trick that can improve stability in practice.
Based on an empirical observation on gradient changes, we freeze the patch projection layer in ViT, \ie, we use fixed random patch projection. We empirically show that this trick alleviates the instability issue in several scenarios and consistently increases accuracy.

We benchmark and ablate self-supervised ViT in a variety of cases. We provide ViT results in several self-supervised frameworks. We conduct ablations on architecture designs and discuss the implications.
Furthermore, we explore scaling up the ViT models, including the non-trivial \mbox{ViT-Large} and \mbox{ViT-Huge} \cite{Dosovitskiy2021} ---
the latter has 40$\times$ more computation than ResNet-50 \cite{He2016}.
Based on these experimental results, we discuss both the currently positive evidence as well as the challenges and open questions. 

We report that self-supervised Transformers can achieve strong results using a contrastive learning framework, compared against masked auto-encoding (Table~\ref{tab:sota_ssl_transformers}). This behavior of Transformers differs from the existing trend in NLP.
Moreover, as a promising signal, our bigger self-supervised ViT can achieve better accuracy, unlike the ImageNet-supervised ViT in \cite{Dosovitskiy2021} whose accuracy degrades if getting bigger.
For instance, for the very big \mbox{ViT-Large}, our self-supervised pre-training can outperform its \mbox{supervised} pre-training counterpart for transfer learning in certain cases. This presents a proof-of-concept scenario where self-supervised pre-training is needed. 

In addition, we report that our self-supervised ViT models have competitive results \vs the \emph{big} convolutional ResNets in prior art \cite{Chen2020b,Grill2020}. On one hand, this comparison shows the potential of ViT, especially considering that it achieves these results using relatively ``\emph{fewer inductive biases}" \cite{Dosovitskiy2021}. On the other hand, we suggest that there could be room for self-supervised ViT models to further improve.
As one example, we observe that \emph{removing the position embedding} in ViT only degrades accuracy by a small margin. This reveals that self-supervised ViT can learn strong representations \emph{without} the positional inductive bias, but it also implies that the positional information has not been sufficiently exploited.

In summary, we believe that the evidence, challenges, and open questions in this study are worth knowing, if self-supervised Transformers will close the gap in pre-training between vision and language. We hope our data points and experience will be useful to push this frontier.

\section{Related Work}

\paragraph{Self-supervised visual representation learning.} In computer vision, contrastive learning \cite{Hadsell2006} has become increasingly successful for self-supervised learning, \eg, \cite{Wu2018a,Oord2018,Hjelm2019,Bachman2019,He2020,Chen2020}. The methodology is to learn representations that attract similar (positive) samples and dispel different (negative) samples. The representations from contrastive self-supervised pre-training can outperform their supervised counterparts in certain tasks \cite{He2020,Chen2020}.

Contrastive learning is commonly instantiated as some forms of Siamese networks \cite{Bromley1994}. Recently, a series of works \cite{Grill2020,Caron2020,Chen2021} retain the Siamese architectures but eliminate the requirement of negative samples.
The success of these methods suggest that it is of central importance to learn invariant features by matching positive samples.

\paragraph{Transformers.} Transformers \cite{Vaswani2017} were originally introduced for machine translation and later became a dominant backbone in NLP \cite{Radford2018,Devlin2019,Radford2019,Brown2020}. 
The long-range, self-attentional behavior makes Transformers an effective tool given the non-local, relational nature of languages.

There have been continuous efforts on generalizing Transformers to computer vision \cite{Wang2018,Cao2019,Ramachandran2019,Zhao2020,Carion2020,Dosovitskiy2021}. The recent work on Vision Transformers (ViT) \cite{Dosovitskiy2021} greatly pushes this frontier.
ViT is purely Transformer-based, rather than interlaced with non-degenerated (\ie, non-1$\times$1) convolutions.\footnotemark~This largely closes the architectural gap between NLP and vision.
ViT achieves compelling accuracy in supervised learning, especially with large-scale data and high-capacity models.
Given these properties, we believe ViT is a must-study baseline for self-supervised learning in computer vision.

\footnotetext{We argue that it is \emph{imprecise} to simply compare self-attention against ``convolutions". Convolutions \cite{LeCun1989} by definition have several properties: weight-sharing, locally-connected, translation-equivariant. All projection layers in a self-attention block have all these properties of convolutions, and are equivalent to 1$\times$1 convolutions. 
The counterpart of self-attention is more appropriately the non-degenerated (\eg, 3$\times$3) convolutions.}

\paragraph{Self-supervised Transformers for vision.} In pioneering works \cite{Chen2020c,Dosovitskiy2021}, training self-supervised Transformers for vision problems in general follows the \emph{masked auto-encoding} paradigm in NLP \cite{Radford2018,Devlin2019} (Table~\ref{tab:sota_ssl_transformers}). 
iGPT \cite{Chen2020c} masks and reconstructs pixels, and the self-supervised variant of ViT in \cite{Dosovitskiy2021} masks and reconstructs patches.
In this work, we focus on training Transformers in the contrastive/Siamese paradigm, in which the loss is not defined for reconstructing the inputs.

\section{MoCo v3}

We introduce a ``\mbox{MoCo~v3}'' framework that facilitates our study.
MoCo v3 is an incremental improvement of MoCo v1/2 \cite{He2020,Chen2020a}, and we strike for a better balance between simplicity, accuracy, and scalability. The pseudocode of MoCo~v3 is in Alg.~\ref{alg:code}, described next. 

As common practice (\eg, \cite{He2020,Chen2020}), we take two crops for each image under random data augmentation. They are encoded by two encoders, $f_q$ and $f_k$, with output vectors $q$ and $k$. Intuitively, $q$ behaves like a ``query'' \cite{He2020}, and the goal of learning is to retrieve the corresponding ``key''. This is formulated as minimizing a contrastive loss function \cite{Hadsell2006}. We adopt the form of InfoNCE \cite{Oord2018}:
\begin{equation}
\small
\mathcal{L}_{q} = -\log \frac{\exp(q{\cdot}k^+ / \tau)}{\exp(q{\cdot}k^+ / \tau) + {\displaystyle\sum_{k^-}}\exp(q{\cdot}k^-  / \tau)}.
\label{eq:infonce}
\end{equation}
Here $k_{+}$ is $f_k$'s output on the same image as $q$, known as $q$'s positive sample. The set $\{k^{-}\}$ consists of $f_k$'s outputs from other images, known as $q$'s negative samples.
$\tau$ is a temperature hyper-parameter \cite{Wu2018a} for $\ell_2$-normalized $q$, $k$.

Following \cite{Ye2019,Hjelm2019,Bachman2019,Chen2020}, in MoCo~v3 we use the keys that naturally co-exist in the same batch. We abandon the memory queue \cite{He2020}, which we find has diminishing gain if the batch is sufficiently large (\eg, 4096).
With this simplification, the contrastive loss in (\ref{eq:infonce}) can be implemented by a few lines of code: see \lstinline{ctr(q, k)} in Alg.~\ref{alg:code}.
We adopt a symmetrized loss \cite{Grill2020,Caron2020,Chen2021}: \lstinline{ctr(q1, k2)+ctr(q2, k1)}.

Our encoder $f_q$ consists of a backbone (\eg, ResNet, ViT), a \emph{projection} head \cite{Chen2020}, and an extra \emph{prediction} head \cite{Grill2020}; the encoder $f_k$ has the backbone and projection head, but not the prediction head. $f_k$ is updated by the moving-average of $f_q$ \cite{He2020}, excluding the prediction head.

As a reference, we examine the MoCo~v3 accuracy with ResNet-50 (R50) (detailed in appendix). This table compares the linear probing accuracy in ImageNet:
\begin{center}
\vspace{-.2em}
\small
\tablestyle{2pt}{1.1}
\begin{tabular}{x{52}|x{52}x{52}x{52}}
R50, 800-ep & MoCo v2 \cite{Chen2020a} & MoCo v2+ \cite{Chen2021} & MoCo v3 \\
\shline
linear acc. & 71.1 & 72.2 & \textbf{73.8} \\
\end{tabular}
\vspace{-.2em}
\end{center}
The improvement here is mainly due to the extra prediction head and large-batch (4096) training.

\begin{algorithm}[t]
\caption{MoCo v3: PyTorch-like Pseudocode}
\label{alg:code}
\algcomment{
\textbf{Notes}: \texttt{mm} is matrix multiplication. \texttt{k.t()} is \texttt{k}'s transpose. The prediction head is excluded from \texttt{f\_k} (and thus the momentum update).
}
\definecolor{codeblue}{rgb}{0.25,0.5,0.5}
\definecolor{codekw}{rgb}{0.85, 0.18, 0.50}
\begin{lstlisting}[language=python]
# f_q: encoder: backbone + proj mlp + pred mlp
# f_k: momentum encoder: backbone + proj mlp
# m: momentum coefficient
# tau: temperature

for x in loader:  # load a minibatch x with N samples
    x1, x2 = aug(x), aug(x)  # augmentation
    q1, q2 = f_q(x1), f_q(x2)  # queries: [N, C] each
    k1, k2 = f_k(x1), f_k(x2)  # keys: [N, C] each

    loss = ctr(q1, k2) + ctr(q2, k1)  # symmetrized
    loss.backward()
    
    update(f_q)  # optimizer update: f_q
    f_k = m*f_k + (1-m)*f_q  # momentum update: f_k
  
# contrastive loss
def ctr(q, k):
    logits = mm(q, k.t())  # [N, N] pairs
    labels = range(N)  # positives are in diagonal
    loss = CrossEntropyLoss(logits/tau, labels) 
    return 2 * tau * loss
\end{lstlisting}
\end{algorithm}

\section{Stability of Self-Supervised ViT Training}

In principle, it is straightforward to replace a ResNet backbone with a ViT backbone in the contrastive/Siamese self-supervised frameworks. But in practice, a main challenge we have met is the \emph{instability} of training. 

We observe that the instability problem can not be simply reflected by accuracy numbers. In fact, as we will show, the training is ``apparently good'' and provides decent results, even when it is potentially unstable.
To reveal the instability, we monitor the kNN curves \cite{Wu2018a} (see appendix) during training. In Sec.~\ref{subsec:empirical}, we study how the basic factors influence stability.
The curves suggest that the training can be ``partially successful", or in other words, ``partially failed". 
In Sec.~\ref{subsec:trick}, we explore a simple trick that can improve stability. As a result, the accuracy is improved in various cases. 

\begin{figure}[t]
\centering
\includegraphics[width=.85\linewidth]{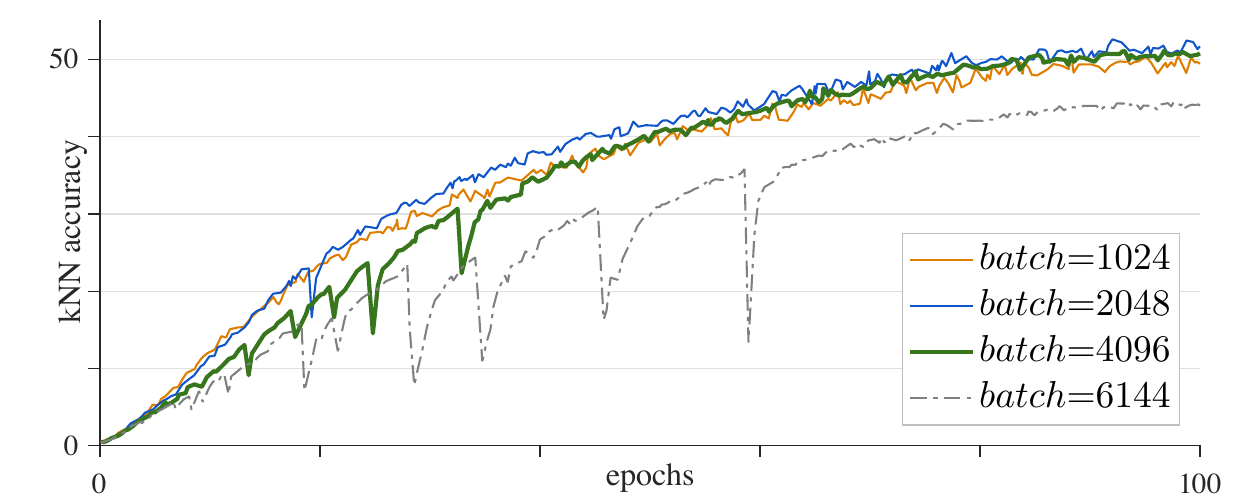}\vspace{.1em}
\tablestyle{4pt}{1.1}
\begin{tabular}{c|x{36}x{36}x{36}x{36}}
batch & 1024 & 2048 & 4096 & 6144 \\
\shline
linear acc. & 71.5 & 72.6 & 72.2 & 69.7 \\ 
\end{tabular}
\vspace{.3em}
\caption{
\textbf{Training curves of different batch sizes} (\mbox{MoCo~v3}, ViT-B/16, 100-epoch ImageNet, AdamW, $lr{=}1.0e\text{-}4$).
\label{fig:curves_batch}
}
\end{figure}

\subsection{Empirical Observations on Basic Factors}\label{subsec:empirical}

\paragraph{Batch size.} ViT models in \cite{Dosovitskiy2021} are by design computationally heavy (see Table~\ref{tab:config} and~\ref{tab:time}), and large-batch training \cite{Goyal2017,You2017,You2020} is a desirable solution to big ViT models. A large batch is also beneficial for accuracy in recent self-supervised learning methods \cite{Chen2020,Grill2020,Caron2020}. 
Fig.~\ref{fig:curves_batch} presents the training curves with different batch sizes. 

A batch of 1k and 2k produces reasonably smooth curves, with 71.5\% and 72.6\% linear probing accuracy. 
In this regime, the larger batch improves accuracy thanks to more negative samples \cite{He2020,Chen2020}.
The curve of a 4k batch becomes noticeably unstable: see the ``dips" in Fig.~\ref{fig:curves_batch}.
It has 72.2\% linear probing accuracy. Although this seems to be a marginal decrease \vs the 2k batch, its accuracy is harmed by the instability, as we will show in the next subsection.

The curve of a 6k batch has worse failure patterns (big dips in Fig.~\ref{fig:curves_batch}).
We hypothesize that the training is partially \emph{restarted} and jumps out of the current local optimum, then seeks a new trajectory. As a consequence, the training does not diverge, but the accuracy depends on how good the local restart is.
When this partial failure happens, it still provides an apparently decent result (69.7\%). This behavior is harmful to explorative research: unlike catastrophic failure that is easily noticeable, the small degradation can be fully hidden.

We also find that the mild instability does \emph{not} result in a noticeably large variation. In many of our ablations, running the same configuration for a second trial often results in a small difference of 0.1$\app$0.3\%. This also makes it difficult to notice the potential degradation caused by instability.

\begin{figure}[t]
\centering
\includegraphics[width=.85\linewidth]{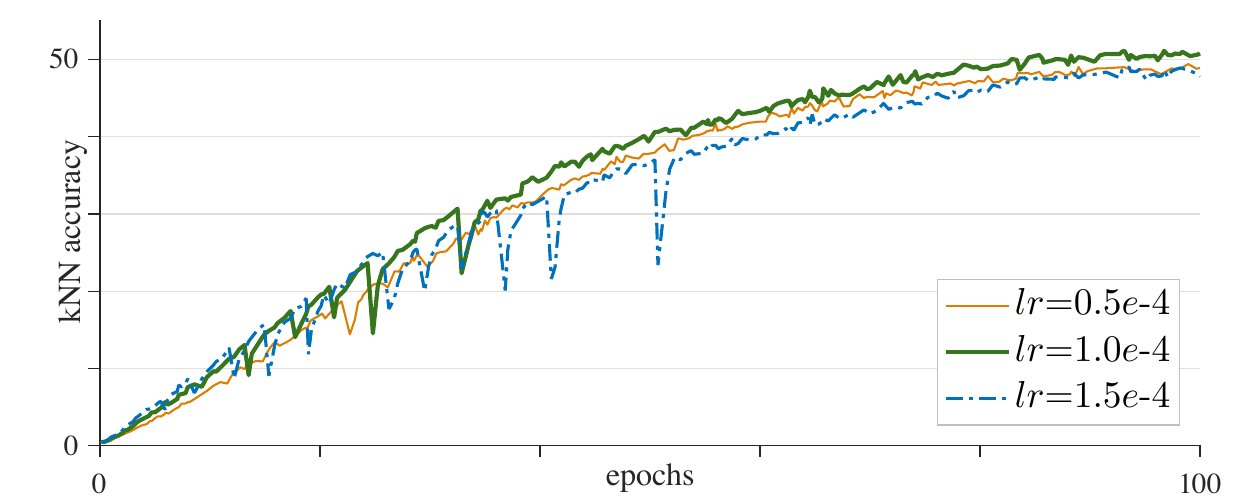}\vspace{.1em}
\tablestyle{4pt}{1.1}
\begin{tabular}{c|x{36}x{36}x{36}}
$lr$, $\times1e\text{-}4$ & 0.5 & 1.0 & 1.5 \\
\shline
linear acc. & 70.4 & 72.2 & 71.7 \\ 
\end{tabular}
\vspace{.3em}
\caption{
\textbf{Training curves of different learning rates} (MoCo v3, ViT-B/16, 100-epoch ImageNet, AdamW, batch 4096).
\label{fig:curves_lr}
}
\end{figure}

\paragraph{Learning rate.} In practice, the learning rate is often scaled when the batch size increases \cite{Krizhevsky2014,Goyal2017}. 
In all experiments in this paper, we adopt the linear scaling rule \cite{Krizhevsky2014,Goyal2017}: we set the learning rate as $lr{\times}$BatchSize${/}256$, where $lr$ is a ``base" learning rate.
$lr$ is the hyper-parameter being set \cite{He2020,Chen2020,Grill2020}.
In Fig.~\ref{fig:curves_lr} we study the influence of $lr$.

When $lr$ is smaller, the training is more stable, but it is prone to under-fitting. In Fig.~\ref{fig:curves_lr}, $lr{=}0.5e\text{-}4$ has 1.8\% worse accuracy than $lr{=}1.0e\text{-}4$ (70.4 \vs 72.2). In this regime, the accuracy is determined by fitting \vs under-fitting.
Training with a larger $lr$ becomes less stable. Fig.~\ref{fig:curves_lr} shows that $lr{=}1.5e\text{-}4$ for this setting has more dips in the curve, and its accuracy is lower.
In this regime, the accuracy is determined by stability.

\begin{figure}[t]
\centering
\includegraphics[width=.85\linewidth]{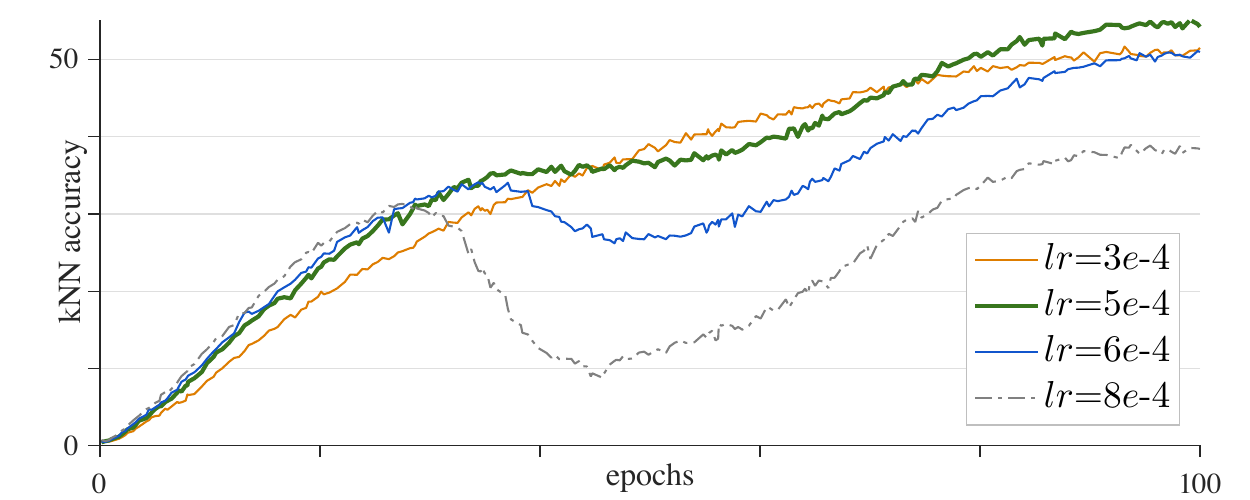}\vspace{.1em}
\tablestyle{4pt}{1.1}
\begin{tabular}{c|x{36}x{36}x{36}x{36}}
\emph{lr}, $\times1e\text{-}4$ & 3.0 & 5.0 & 6.0 & 8.0 \\
\shline
linear acc. & 71.6 & 72.5 & 70.9 & 66.5 \\ 
\end{tabular}
\vspace{.3em}
\caption{
\textbf{Training curves of LAMB optimizer} (MoCo v3, ViT-B/16, 100-epoch ImageNet, $wd{=}1e\text{-}3$, batch 4096).
\label{fig:curves_lamb}
}
\end{figure}

\paragraph{Optimizer.} By default, we use \mbox{AdamW} \cite{Loshchilov2019} as the optimizer, which is the common choice for training ViT models \cite{Dosovitskiy2021,Touvron2020,Radford2021}.\footnotemark~On the other hand, recent self-supervised methods \cite{Chen2020,Grill2020,Caron2020} are based on the LARS optimizer \cite{You2017} for large-batch training.
In Fig.~\ref{fig:curves_lamb}, we study the LAMB optimizer \cite{You2020}, which is an AdamW-counterpart of LARS.

\footnotetext{In original ViT \cite{Dosovitskiy2021} in JAX, the weight decay is ``AdamW style'':
{\fontsize{6.2pt}{1em}\selectfont\url{https://github.com/google/flax/blob/master/flax/optim/adam.py}}
} 

Given an appropriate learning rate ($lr{=}5e\text{-}4$, Fig.~\ref{fig:curves_lamb}), LAMB achieves slightly better accuracy (72.5\%) than AdamW.
But the accuracy drops rapidly when $lr$ is larger than the optimal value. LAMB with $lr{=}6e\text{-}4$ and $8e\text{-}4$ has 1.6\% and 6.0\% lower accuracy.
Interestingly, the training curves are still smooth, but they degrade gradually in the middle. 
We hypothesize that although LAMB can avoid sudden change in the gradients, the negative impact of unreliable gradients is accumulated.

During our exploration, we find that LAMB can achieve comparable accuracy with AdamW, if $lr$ is appropriately chosen.
But the sensitivity to $lr$ makes it difficult to ablate different architecture designs without extra $lr$ search. As a result, we opt to use AdamW in other parts of this paper.

\begin{figure}[t]
\centering
\includegraphics[width=.99\linewidth]{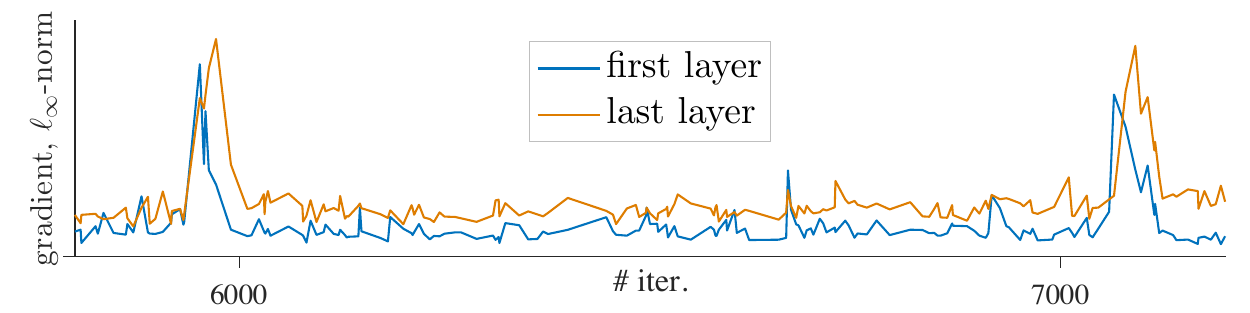}\vspace{.1em}
\caption{
We monitor the gradient magnitude, shown as relative values for the layer. A ``spike'' in the gradient causes a ``dip'' in the training curve. We observe that a spike happens earlier in the first layer, and are delayed by tens of iterations in the last layers.
\label{fig:grad_monitor}
}
\end{figure}

\subsection{A Trick for Improving Stability}\label{subsec:trick}

All these experiments suggest that instability is a major issue.
Next we describe a simple trick that can improve stability in various cases in our experiments.

During training, we notice that a sudden change of gradients (a ``spike'' in Fig.~\ref{fig:grad_monitor}) causes a ``dip'' in the training curve, which is as expected.
By comparing all layers' gradients, we observe that the gradient spikes happen earlier in the first layer (patch projection), and are delayed by couples of iterations in the last layers (see Fig.~\ref{fig:grad_monitor}). Based on this observation, we hypothesize that the instability happens earlier in the shallower layers.
Motivated by this, we explore freezing the patch projection layer during training. In other words, we use a fixed \textbf{\emph{random patch projection}} layer to embed the patches, which is not learned. This can be easily done by applying a stop-gradient operation right after this layer.

\begin{figure}[t]
\centering
\includegraphics[width=.85\linewidth]{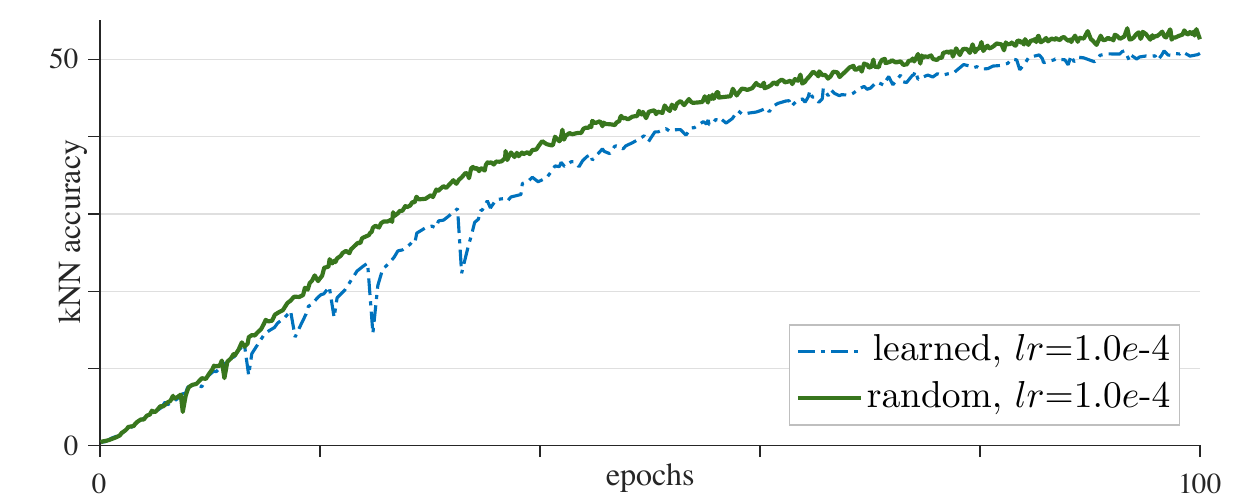}
\includegraphics[width=.85\linewidth]{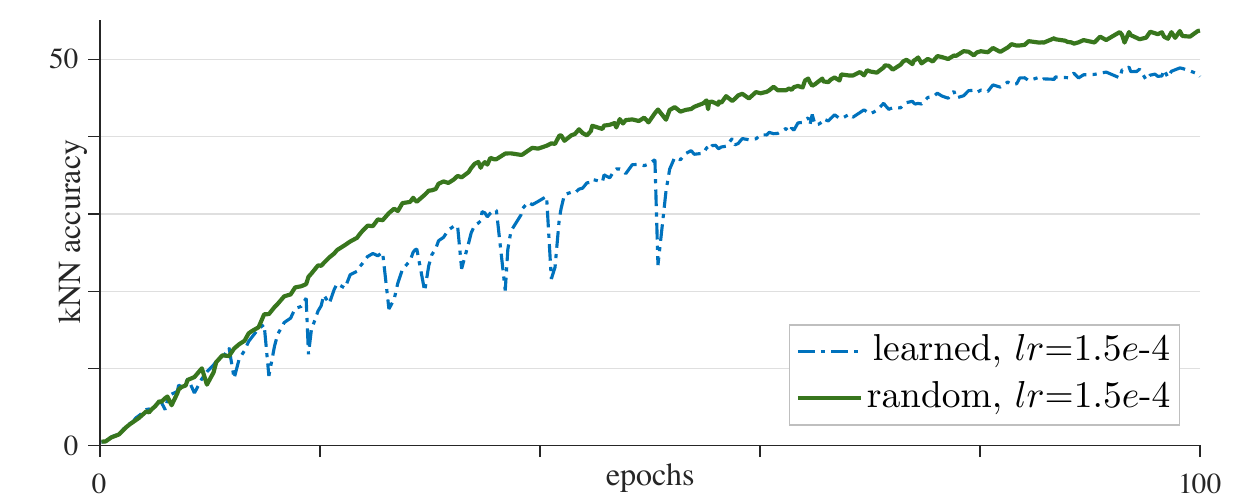}
\vspace{.3em}
\tablestyle{4pt}{1.1}
\begin{tabular}{c|x{36}x{36}x{36}}
$lr$, $\times10^{-4}$ & 0.5 & 1.0 & 1.5 \\
\shline
learned patch proj. & 70.4 & 72.2 & 71.7 \\
random patch proj. & \textbf{70.8} & \textbf{72.8} & \textbf{73.4}
\end{tabular}
\vspace{.3em}
\caption{
\textbf{Random \vs learned patch projection} (MoCo v3, ViT-B/16, 100-epoch ImageNet, AdamW, batch 4096). \textbf{Top}: $lr{=}1.0e\text{-}4$. \textbf{Bottom}: $lr{=}1.5e\text{-}4$.
\label{fig:curves_stopgrad}
}
\end{figure}
\begin{figure}[t]
\centering
\includegraphics[width=.85\linewidth]{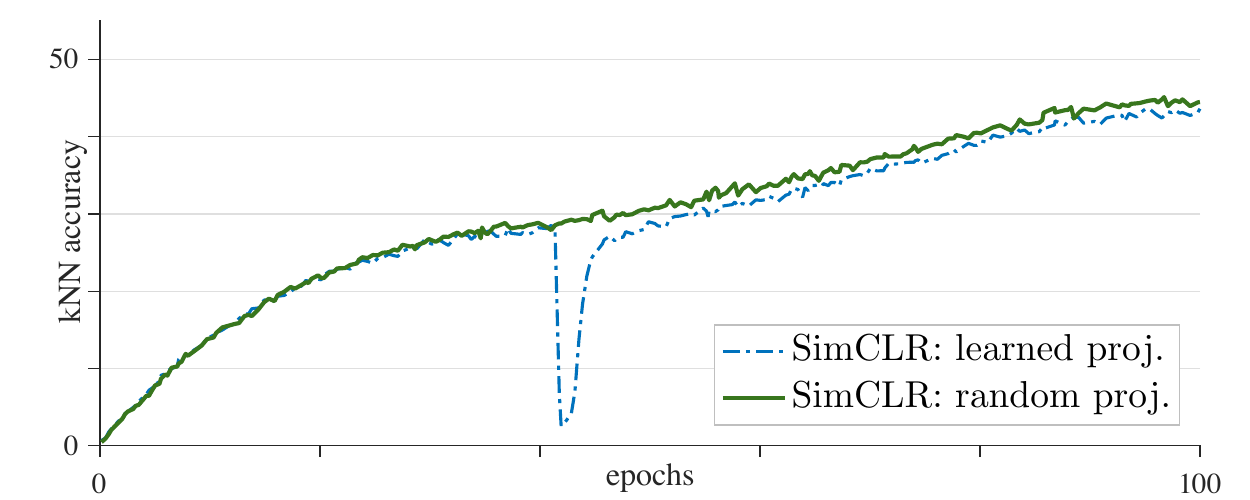}
\includegraphics[width=.85\linewidth]{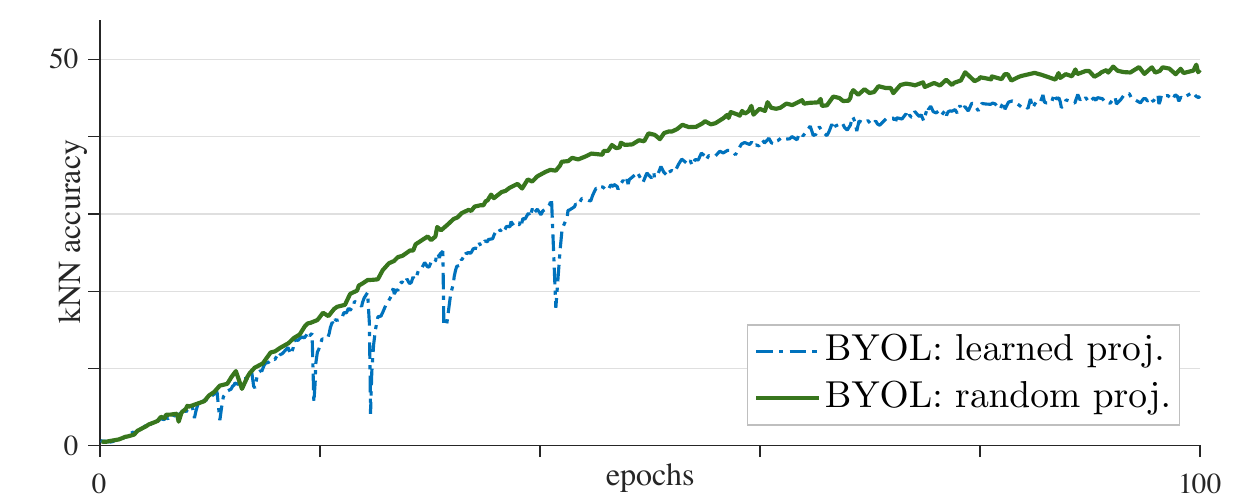}
\vspace{.3em}
\tablestyle{4pt}{1.1}
\begin{tabular}{c|x{36}x{36}}
& SimCLR & BYOL \\ 
\shline
learned patch proj. & 69.3 & 69.7 \\ 
random patch proj. & \textbf{70.1} & \textbf{71.0} 
\end{tabular}
\vspace{.3em}
\caption{
\textbf{Random \vs learned patch projection} (ViT-B/16, 100-epoch ImageNet, AdamW, batch 4096).
\textbf{Top}: SimCLR: $lr{=}2e\text{-}4$, $wd{=}0.1$. \textbf{Bottom}: BYOL: $lr{=}1e\text{-}4$, $wd{=}0.03$.
\label{fig:curves_stopgrad_simclr}
}
\vspace{-1em}
\end{figure}

\paragraph{Comparisons.}
In Fig.~\ref{fig:curves_stopgrad} we show the MoCo v3 results with learnable \vs random patch projection. Random patch projection stabilizes training, with smoother and better training curves. This stability benefits the final accuracy, boosting the accuracy by \textbf{1.7\%} to 73.4\% at $lr{=}1.5e\text{-}4$.
The improvement is bigger for a larger $lr$ (0.4\%, 0.6\%, 1.7\%). 
This comparison confirms that the training instability is a main issue that impacts accuracy.

\hspace{-.35em}Besides MoCo, we find that other related methods \cite{Chen2020,Grill2020,Caron2020} can also be unstable. Fig.~\ref{fig:curves_stopgrad_simclr} presents the training curves of ViT in SimCLR \cite{Chen2020} and BYOL \cite{Grill2020}. Random patch projection improves stability in both SimCLR and BYOL, and increases the accuracy by 0.8\% and 1.3\%. We also observe the instability issue for SwAV \cite{Caron2020}, in which, however, the loss diverges (NaN) when it is unstable. Random patch projection helps SwAV by enabling a relatively larger $lr$ without diverging, and improves its accuracy from 65.8\% to 66.4\% when using the largest stable $lr$. In sum, this trick is effective in all these self-supervised frameworks.

We have also tried BatchNorm (BN) \cite{Ioffe2015}, WeightNorm (WN) \cite{Salimans2016}, or gradient clip on patch projection. We observe that BN or WN on the learnable patch projection layer does not improve instability, and produces similar results; gradient clip on this layer is useful if given a sufficiently small threshold, which to the extreme becomes freezing the layer.

\paragraph{Discussions.}
It is an interesting observation that it is not necessary to train the patch projection layer. For the standard ViT patch size, the patch projection matrix is complete (768-d output for a 3-channel 16$\times$16 patch) or over-complete. 
In this case, random projection should be sufficient to preserve the information of the original patches. 

We note that freezing the first layer does not change the architecture, and it actually narrows down the solution space. This indicates that the underlying problem is on optimization. 
The trick alleviates the issue, but does not solve it. The model can still be unstable if $lr$ is too big.
The first layer is unlikely the essential reason for the instability; instead, the issue concerns all layers. The first layer is merely easier to be handled separately, \eg, it is the only non-Transformer layer in the backbone. We hope to see a more fundamental solution in future work.

\section{Implementation Details}

This section describes the details of ViT+MoCo~v3. More subtleties are described in the appendix.

\paragraph{Optimizer.} By default we use AdamW \cite{Loshchilov2019} and a batch size of 4096 \cite{Chen2020,Grill2020,Caron2020}. 
We search for $lr$ and weight decay $wd$ based on 100-epoch results, and then apply it for longer training. We adopt learning rate warmup \cite{Goyal2017} for 40 epochs (as per ``\emph{warmup of 10k steps}'', Table~4 in \cite{Dosovitskiy2021}). 
This long warmup helps alleviate instability, though all unstable results are already with this warmup. 
After warmup, $lr$ follows a cosine decay schedule \cite{Loshchilov2016}.

\paragraph{MLP heads.} The projection head \cite{Chen2020} is a 3-layer MLP, following \cite{Chen2020b}. The prediction head \cite{Grill2020} is a 2-layer MLP. The hidden layers of both MLPs are 4096-d and are with ReLU \cite{Nair2010}; the output layers of both MLPs are 256-d, \mbox{without} ReLU. In MoCo v3, all layers in both MLPs have BN \cite{Ioffe2017}, following SimCLR \cite{Chen2020}. The MLP heads of BYOL/SwAV have different BN designs (see appendix).

\paragraph{Loss.} We scale the contrastive loss in (\ref{eq:infonce}) by a constant $2\tau$ (see Alg.~\ref{alg:code}), following \cite{Grill2020}'s appendix. This scale is redundant because it can be absorbed by adjusting $lr$ and $wd$. But this scale makes it less sensitive to the $\tau$ value when $lr$ and $wd$ are fixed. We set $\tau{=}0.2$ \cite{Chen2020a} as the default.

\paragraph{ViT architecture.} We closely follow the designs in \cite{Dosovitskiy2021}. The input patch size is 16$\times$16 or 14$\times$14 (`/16' or `/14'), and after projection it results in a sequence of length 196 or 256 for a 224$\times$224 input. 
Position embeddings are added to the sequence, and we use the sine-cosine variant \cite{Vaswani2017} in 2-D.
This sequence is concatenated with a learnable class token.
The sequence is then encoded by a stack of Transformer blocks \cite{Vaswani2017} following the design in \cite{Dosovitskiy2021}.
The class token after the last block (and after the final LayerNorm \cite{Ba2016}) is treated as the output of the backbone, and is the input to the MLP heads.

\paragraph{Linear probing.} Following common practice, we evaluate the representation quality by linear probing. After self-supervised pre-training, we remove the MLP heads and train a supervised linear classifier on frozen features. We use the SGD optimizer, with a batch size of 4096, $wd$ of 0, and sweep $lr$ for each case.
We train this supervised classifier for 90 epochs in the ImageNet training set, using only random resized cropping and flipping augmentation. We evaluate single-crop top-1 accuracy in the validation set.

\begin{table}[t]
\centering
\tablestyle{6pt}{1.1}
\begin{tabular}{l|crcrc}
model & blocks & dim & heads & params \\
\shline
ViT-Small & 12 & 384 & 12 & 22 M \\
ViT-Base \cite{Dosovitskiy2021} & 12 & 768 & 12 & 86 M \\
ViT-Large \cite{Dosovitskiy2021} & 24 & 1024 & 16 & 304 M \\
ViT-Huge \cite{Dosovitskiy2021} & 32 & 1280 & 16 & 632 M \\
\end{tabular}
\vspace{.5em}
\caption{\textbf{Configurations of ViT models} in our experiments. Here ``blocks'' is the number of Transformer blocks, ``dim'' is the input/output channel dimension of all blocks, and ``heads'' is the number of heads in multi-head attention. The MLP hidden dimension is 4$\times$dim.
\label{tab:config}
}
\end{table}

\begin{table}[t]
\centering
\tablestyle{5pt}{1.1}
\begin{tabular}{x{48}|rrrrr}
model &
FLOPs &
\vs R50 & TPUs & hours \\ 
\shline
ViT-S/16 & 4.6 G & 1.1$\times$ & 256 & 1.2 \\ 
ViT-B/16 & 17.5 G & 4.3$\times$ & 256 & 2.1 \\ 
ViT-L/16 & 61.3 G & 15.0$\times$ & 256 & 6.1 \\ 
ViT-H/14 & 166.7 G & 40.7$\times$ & 512 & 9.8 \\ 
\end{tabular}
\vspace{.5em}
\caption{\textbf{Training time of ViT + MoCo v3}, per 100 ImageNet-epochs, in our TensorFlow implementation.
The FLOPs number (in multiply-adds) is per 224$\times$224 crop, and ``\vs R50'' is the relative FLOPs \vs ResNet-50 (4.1G).
\label{tab:time}
}
\vspace{-1em}
\end{table}

\section{Experimental Results}\label{sec:exp}

In this section we benchmark and ablate self-supervised ViT.
We perform self-supervised training on the 1.28M ImageNet training set \cite{Deng2009}, and evaluate by linear probing.

Table~\ref{tab:config} summarizes the ViT configurations we study. \mbox{ViT-B/L/H} follow \cite{Dosovitskiy2021}, and ViT-S is similar to that in \cite{Touvron2020}. 
We use ViT-B by default in our ablations.

\paragraph{Training time.} We train our models in TPUs (v3) that are publicly available in Google Cloud Platform (GCP). Table~\ref{tab:time} summarizes the training time (per 100 epochs).
It takes 2.1 hours training ViT-B for 100 epochs, and our ablations typically take 6.3 hours each (300 epochs). This is a competitive performance, as it enables us to ablate many design decisions. The TPU implemenataion also makes it possible to explore the ViT-H model, which takes 9.8 hours per 100 epochs using 512 TPUs. This is a gigantic scale of training: for the 300-epoch ViT-H, this amounts to $\app$625 TPU$\cdot${days}, or $\app$1.7 TPU$\cdot${years} of training.

We also verify our models in GPUs using PyTorch. It takes 24 hours for ViT-B in 128 GPUs (\vs 2.1 hours in 256 TPUs). 
With an increasing number of devices, we observe that TPUs scale up more favorably than GPUs. While further engineering optimization could speed up our GPU system, we opt to use the TPU system for the ease of research.

\subsection{Self-supervised Learning Frameworks}

We benchmark self-supervised ViT in four frameworks: MoCo v3, SimCLR \cite{Chen2020}, BYOL \cite{Grill2020}, and SwAV \cite{Caron2020}.\hspace{-.3em}
We use the same random projection trick in all cases. We sweep $lr$ and $wd$ for each individual framework for fair comparisons.

\begin{figure}[t]
\centering
\begin{minipage}[c]{0.42\linewidth}
\hspace{.5em}
\includegraphics[width=.8\linewidth]{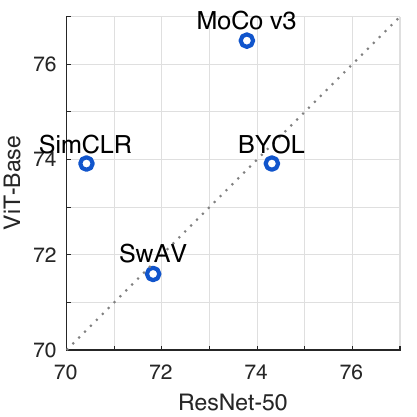}
\end{minipage}
\begin{minipage}[c]{0.57\linewidth}
\caption{
Different self-supervised learning frameworks perform differently between \mbox{R-50} \cite{He2016} (x-axis) and ViT-B \cite{Dosovitskiy2021} (y-axis). 
The numbers are ImageNet linear probing accuracy from Table~\ref{tab:vit_vs_frameworks}.
\label{fig:vit_vs_r50}
}
\vspace{1em}
\end{minipage}
\\\vspace{-1em}
\end{figure}

Table~\ref{tab:vit_vs_frameworks} reports the results of ViT-S/16 and ViT-B/16. MoCo v3 has better accuracy on ViT than other frameworks.
The \emph{relative} accuracy among these methods is different between ViT-B and R50: see Fig~\ref{fig:vit_vs_r50}. MoCo~v3 and SimCLR are more favorable for ViT-B than R50 (above the diagonal line).

\begin{table}[t]
\centering
\tablestyle{4pt}{1.1}
\begin{tabular}{l|x{32}x{32}x{32}x{32}}
model & MoCo v3 & SimCLR & BYOL & SwAV \\
\shline
\demph{R-50, 800-ep} & \demph{73.8} & \demph{70.4} & \demph{\textbf{74.3}} & \demph{71.8} \\
ViT-S, 300-ep & \textbf{72.5} & 69.0 & 71.0 & 67.1 \\
ViT-B, 300-ep & \textbf{76.5} & 73.9 & 73.9 & 71.6 \\
\end{tabular}
\vspace{.5em}
\caption{\textbf{ViT-S/16 and ViT-B/16 in different self-supervised learning frameworks} (ImageNet, linear probing).
R-50 results of other frameworks are from the improved implementation in \cite{Chen2021}. For fair comparisons, all are pre-trained with two 224$\times$224 crops for each image (multi-crop training \cite{Caron2020} could improve results, which is beyond the focus of this work).
\label{tab:vit_vs_frameworks}
}
\end{table}

\subsection{Ablations of ViT + MoCo v3}

Next we ablate the designs of the ViT + MoCo v3 system. We use random patch projection in all ablations.

\paragraph{Position embedding.} The following table compares the choice of position embedding (our default is sin-cos):
\begin{center}
\vspace{-.3em}
\small
\tablestyle{2pt}{1.1}
\begin{tabular}{x{52}|x{48}x{48}x{48}}
ViT-B, 300-ep & sin-cos & learned & none \\
\shline
linear acc. & {76.5} & 76.1 & 74.9 \\
\end{tabular}
\vspace{-.3em}
\end{center}
The learned version works well, but not better than sin-cos.
Surprisingly, the model works decently even with \emph{no} position embedding (74.9\%). The capability to encode positions contributes only 1.6\%. We believe this data point reveals both strengths and limitations of the current model. On the positive side, it suggests that the model can learn strong representations just by a \emph{set} of patches, which are fully \emph{permutation-invariant}. This is analogous to bag-of-words models \cite{Sivic2003}. This model has \emph{no} positional inductive bias.
On the negative side, it also suggests that the model has not made good use of positions, and the gesture of the object contributes relatively little to the representation. We hope this data point will draw attention to future study.

\paragraph{Class token.}
The following table ablates the role of the class token [\texttt{CLS}] in ViT:
\begin{center}
\vspace{-.3em}
\small
\tablestyle{6pt}{1.1}
\begin{tabular}{c|ccc}
ViT-B, 300-ep & w/ [\texttt{CLS}] & w/o [\texttt{CLS}]; LN+pool  & w/o [\texttt{CLS}]; pool \\
\shline
linear acc. & {76.5} & 69.7 & 76.3 \\
\end{tabular}
\vspace{-.3em}
\end{center}
Global average pooling is used right after the final block if [\texttt{CLS}] is not used. ViT has an extra LayerNorm (LN) after the final block \cite{Dosovitskiy2021}, and if we keep this LN and remove [\texttt{CLS}], the result is much worse (69.7\%). But if we remove this LN and [\texttt{CLS}], the result is nearly unchanged (76.3\%). This comparison indicates that the class token is not essential for the system to work. It also suggests that the choice of normalization layers can make a difference.

\paragraph{BatchNorm in MLP heads.} Unlike the standard ResNets \cite{He2016}, ViT models by default have no BN, and thus all BN layers are in the MLP heads. The following table compares with \vs without BN in the heads:
\begin{center}
\vspace{-.3em}
\small
\tablestyle{2pt}{1.1}
\begin{tabular}{x{56}|x{64}x{64}}
ViT-B, 300-ep & heads w/ BN & heads w/o BN \\
\shline
linear acc. & {76.5} & 74.4 \\
\end{tabular}
\vspace{-.3em}
\end{center}
We have to set the batch size as 2048 when removing BN, otherwise it does not converge.
Removing BN reduces accuracy by 2.1\%. Despite the decrease, this is a completely \emph{BN-free} system. This data point suggests that BN is not necessary for contrastive learning to work, yet appropriate usage of BN can improve accuracy.

\paragraph{Prediction head.} MoCo v3 uses a prediction MLP head as per \cite{Grill2020}. The next table ablates this design:
\begin{center}
\vspace{-.3em}
\small
\tablestyle{2pt}{1.1}
\begin{tabular}{x{56}|x{64}x{64}}
ViT-B, 300-ep & w/ pred. MLP & w/o pred. MLP \\
\shline
linear acc. & {76.5} & 75.5 \\
\end{tabular}
\vspace{-.3em}
\end{center}
Removing the prediction MLP head has a decent result of 75.5\%. While this extra head boosts accuracy, MoCo as a contrastive method does not need the predictor MLP to work, unlike the negative-free methods in \cite{Grill2020,Chen2021}.

\paragraph{Momentum encoder.}
The following table compares the momentum coefficient ($m$) in the momentum encoder:
\begin{center}
\vspace{-.3em}
\small
\tablestyle{2pt}{1.1}
\begin{tabular}{x{56}|x{36}x{36}x{36}x{36}}
ViT-B, 300-ep & $m{=}0$ & $m{=}0.9$ & $m{=}0.99$ & $m{=}0.999$ \\
\shline
linear acc. & 74.3 & 75.6 & {76.5} & 75.0 \\
\end{tabular}
\vspace{-.3em}
\end{center}
The optimal value is $m{=}0.99$ (our default). The case of $m{=}0$ is analogous to SimCLR (plus the prediction head and stop-gradient on the keys), and its accuracy of 74.3\% is similar to SimCLR's (73.9\%, Table~\ref{tab:vit_vs_frameworks}). The usage of the momentum encoder leads to 2.2\% increase.

\paragraph{Training length.} In the following table we report ViT-S/B + MoCo v3 \vs training length:
\begin{center}
\vspace{-.3em}
\small
\tablestyle{4pt}{1.1}
\begin{tabular}{x{48}|x{36}x{36}}
& 300-ep & 600-ep \\
\shline
ViT-S/16 & 72.5 & {73.4} \\
ViT-B/16 & 76.5 & {76.7} \\
\end{tabular}
\vspace{-.3em}
\end{center}
The smaller ViT-S enjoys the benefit of training longer, and improves by 0.9\% when extending to 600 epochs. This is similar to the behavior of R50, which was typically trained for 800 epochs \cite{Chen2020}.
But the gain of training longer is diminishing on ViT-B. Based on this ablation, we train the bigger ViT-L/H for 300 epochs presented next (Table~\ref{tab:sota_ssl_transformers}).

\subsection{Comparisons with Prior Art}

\begin{figure}[t]
\vspace{-1em}
\centering
\includegraphics[width=.96\linewidth]{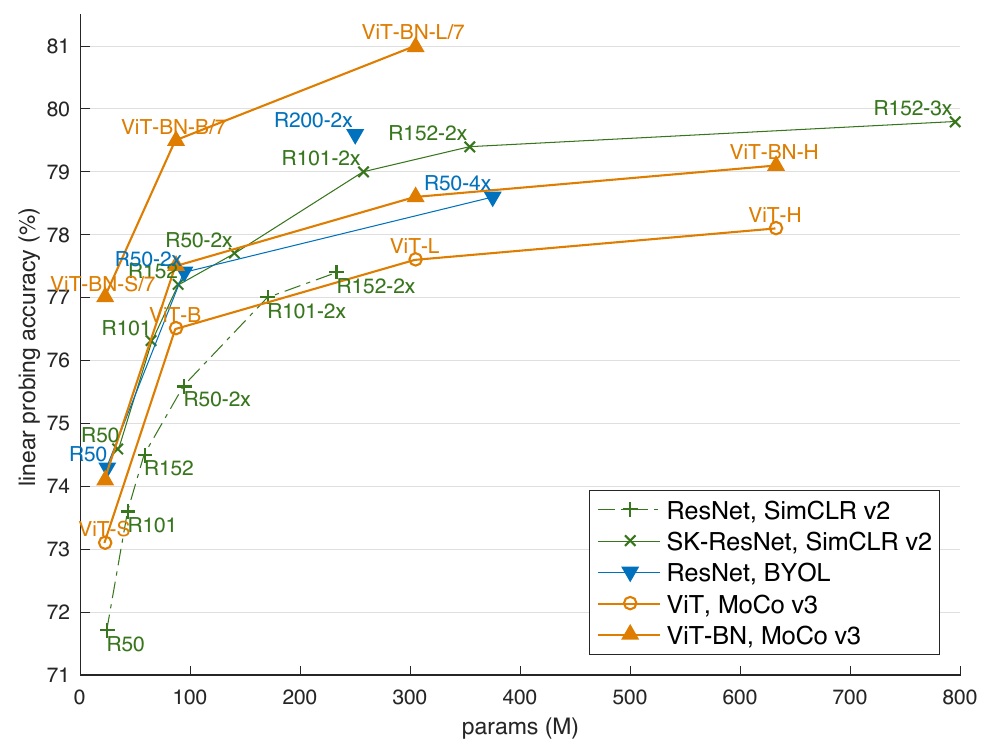}
\tablestyle{4pt}{1.02}
\begin{tabular}{l|x{36}x{36}x{36}x{36}}
MoCo~v3 w/ & ViT-S & ViT-B & ViT-L & ViT-H \\
\shline
ViT baseline & 73.4 & 76.7 & 77.6 & 78.1 \\ 
ViT-BN & 74.1 & 77.5 & 78.6 & 79.1 \\ 
ViT-BN/7 & 77.0 & 79.5 & \textbf{81.0} & - \\ 
\end{tabular}
\vspace{.3em}
\caption{
\textbf{Comparisons with state-of-the-art big ResNets}, presented as parameters-\vs-accuracy trade-off.
All entries are pre-trained with two 224$\times$224 crops, and are evaluated by linear probing. SimCLR v2 results are from Table~1 in \cite{Chen2020b}, and BYOL results are from Table~1 in \cite{Grill2020}.
\label{fig:scatter}
}
\vspace{-1em}
\end{figure}

\begin{table*}[t]
\renewcommand\thetable{6}  
\vspace{-1.5em}
\centering
\tablestyle{1.5pt}{1.05}
\definecolor{green}{HTML}{39b54a}  
\definecolor{red}{HTML}{ea4335}  
\newcommand{\hlg}[1]{\textcolor{green}{#1}}
\newcommand{\hlr}[1]{\textcolor{red}{#1}}
\newcolumntype{z}{>{\centering\arraybackslash}p{31pt}}
\newcolumntype{y}{>{\raggedright\arraybackslash}p{68pt}}
\newcolumntype{d}[1]{>{\raggedright\arraybackslash}p{#1pt}}
\newcolumntype{b}[1]{>{\raggedleft\arraybackslash}p{#1pt}}
\newcommand{\better}[2]{\tablestyle{1pt}{1}
\begin{tabular}{b{16}d{16}}
{#1} &
{\fontsize{7pt}{1em}\selectfont \hlg{$\uparrow$#2}}
\end{tabular}
}
\newcommand{\worse}[2]{\tablestyle{1pt}{1}
\begin{tabular}{b{16}d{16}}
{#1} &
{\fontsize{7pt}{1em}\selectfont \hlr{$\downarrow$#2}}
\end{tabular}
}
\newcommand{\void}[1]{\tablestyle{1pt}{1}
\begin{tabular}{b{16}d{16}}
{#1} &
~
\end{tabular}
}
\begin{tabular}{l|zzz|zzz|zzz|zzz}
& \multicolumn{3}{c|}{CIFAR-10 \cite{Krizhevsky2009}}
& \multicolumn{3}{c|}{CIFAR-100 \cite{Krizhevsky2009}}
& \multicolumn{3}{c|}{Oxford Flowers-102 \cite{Nilsback2008}}
& \multicolumn{3}{c}{Oxford-IIIT-Pets \cite{Parkhi2012}}
\\
pre-train
 & ViT-B & ViT-L & ViT-H
 & ViT-B & ViT-L & ViT-H
 & ViT-B & ViT-L & ViT-H
 & ViT-B & ViT-L & ViT-H \\
\shline
{random init.}
& \void{{77.8}} & \void{{77.1}} & {{75.9}} 
& \void{{48.5}} & \void{{48.3}} & {{48.0}} 
& \void{{54.4}} & \void{{54.3}} & {{52.8}} 
& \void{{40.1}} & \void{{42.8}} & {{40.4}} 
\\
ImNet supervised \cite{Dosovitskiy2021}
& \void{98.1} & \void{97.9} & {n/a} 
& \void{87.1} & \void{86.4} & {n/a} 
& \void{89.5} & \void{89.7} & {n/a} 
& \void{93.8} & \void{93.6} & {n/a}
\\
ImNet self-sup., MoCo v3
& \better{98.9}{0.8} & \better{99.1}{1.2} & {99.1} 
& \better{90.5}{3.4} & \better{91.1}{4.7} & {91.2} 
& \better{97.7}{8.2} & \better{98.6}{8.9} & {98.8} 
& \worse{93.2}{0.6} & \better{93.7}{0.1} & {94.2} 
\end{tabular}
\vspace{.1em}
\caption{
\textbf{Transfer learning} accuracy (\%) in four datasets. All entries are end-to-end fine-tuned \cite{Dosovitskiy2021}. Pre-training are performed in the ImageNet-1k training set.
The models are ViT-B/16, ViT-L/16, and ViT-H/14. Results of ImageNet-supervised pre-training are from Table~3 in \cite{Dosovitskiy2021}. The arrows indicate the changes \wrt the ImageNet-supervised counterparts.
\label{tab:transfer}
}
\vspace{-1em}
\end{table*}

\paragraph{Self-supervised Transformers.}

Table~\ref{tab:sota_ssl_transformers} in Sec.~\ref{sec:intro} presents MoCo~v3 results with different ViT models, compared with state-of-the-art \emph{self-supervised Transformers}. Both iGPT \cite{Chen2020c} and the masked patch prediction in \cite{Dosovitskiy2021} can be categorized as the \emph{masked auto-encoding} paradigm (\eg, GPT \cite{Radford2018} and BERT \cite{Devlin2019}).
Our MoCo-based ViT has higher accuracy and smaller models than iGPT, under the same linear probing protocol and training data. The mask patch prediction in \cite{Dosovitskiy2021} is pre-trained on JFT-300M and end-to-end fine-tuned in ImageNet, which we append as a reference.

Our self-supervised ViT models have \emph{higher} accuracy when the models are \emph{bigger}. 
This is in contrast to the \emph{\mbox{supervised}} results in \cite{Dosovitskiy2021}, where ViT-L has lower accuracy than ViT-B when pre-trained in ImageNet-1k/21k. 
Actually, for ViT-L, our self-supervised pre-training with linear probing (77.6\%) is \emph{better} than the supervised counterpart in \cite{Dosovitskiy2021} (76.53\%) when trained in ImageNet-1k.\footnotemark~These comparisons suggest that self-supervised learning as a tool for generic representation learning is less prone to over-fitting. 

\footnotetext{Stronger regularization could reduce over-fitting for supervised ViT \cite{Touvron2020}, though regularizing the very big ViT-L/H is yet to be explored.}

\paragraph{Comparisons with big ResNets.}\hspace{-.5em}\footnotemark~In Fig.~\ref{fig:scatter} we compare with the state-of-the-art \emph{big} ResNets reported by \mbox{SimCLR~v2} \cite{Chen2020b} and BYOL \cite{Grill2020}. We note that both \mbox{SimCLR~v2} and BYOL use a momentum encoder. 
Our baseline ViT MoCo (the curve of ``ViT, MoCo v3'') is slightly better than ResNet \mbox{SimCLR~v2} in the small-model regime, but the envelopes become just comparable for larger models. \mbox{SimCLR~v2} with SK-ResNet (Selective Kernel \cite{Li2019a}, a form of attention) has a higher envelope. BYOL also has a higher envelope with wider ResNets (1-4$\times$), and has an outstanding point with a deeper ResNet (R200-2$\times$).

\footnotetext{Transformers \cite{Vaswani2017} by design consist of residual blocks \cite{He2016}, and thus are a form of residual networks. In the literature on ``Transformer \vs ResNet", precisely speaking, the term of ``ResNet" refers to the specific design that has non-degenerated (\eg, 3$\times$3) convolutions.}

We notice that this comparison concerns a composition of many choices. As one example, the default ViT backbone in \cite{Dosovitskiy2021} uses {LayerNorm} (LN),
while the default ResNet \cite{He2016} uses {BatchNorm} (BN). These design choices can lead to a systematic gap. In our preliminary experiments, we explore replacing LN with BN in the ViT backbone's MLP blocks (\ie, excluding self-attention blocks).\footnotemark~We simply refer to this as a ``\mbox{ViT-BN}'' backbone.
It leads to $\app$1\% improvement consistently (see Fig.~\ref{fig:scatter}).

\footnotetext{We are trying to replace every LN with BN in the ViT backbone. In preliminary experiments, doing so leads to convergence problems.}

In iGPT \cite{Chen2020c}, accuracy can be improved by using \emph{longer} sequences in the pixel domain.
Here we explore longer sequences by reducing the patch size to 7$\times$7 (``/7'' in Fig.~\ref{fig:scatter}). 
This keeps the model size unchanged, but increases FLOPs to $\app$6$\times$. It can improve the accuracy by $\app$2-3\%. 
The gain of using small patches is also observed in \cite{Caron2021}.
MoCo~v3 achieves \textbf{81.0\%} with \mbox{ViT-BN-L/7}.\footnotemark~As a comparison, the previous best results under the linear probing protocol are 79.8\% with \mbox{SimCLR~v2} (SK-ResNet152-3$\times$), and 79.6\% with BYOL (ResNet200-2$\times$).

\footnotetext{ViT-BN-H/7 is out of memory in our unoptimized implementation.}

\emph{Discussion.} While the bigger self-supervised ViT can achieve better accuracy, the results are saturated.
This is unlike the trend in NLP, where bigger Transformers learn better representations (\eg, \cite{Brown2020}). A potential solution is to use more data. 
The saturation can also be caused by the limited power of the existing \emph{instance-based} pretext task \cite{Wu2018a}. It may be desired to design more difficult pretext tasks.

Our self-supervised ViT models are competitive with the big convolutional ResNets.
It suggests that ViT can learn strong representations with ``\emph{fewer inductive biases}" \cite{Dosovitskiy2021}.
However, we also find that the accuracy only decreases by a bit even if removing the only positional inductive bias (position embedding), suggesting that in our method ViT relies less on positional information than convolutional networks.

\begin{table}[t]
\renewcommand\thetable{5}  
\centering
\tablestyle{6pt}{1.05}
\begin{tabular}{l|c|ccc}
case & pre-train & ViT-S & ViT-B & ViT-L \\
\shline
{masked patch pred. \cite{Dosovitskiy2021}} & JFT-300M & - & 79.9 & - \\
DeiT \cite{Touvron2020} & - & 79.9 & 81.8 & n/a \\
MoCo v3 & ImageNet-1k & \textbf{81.4} & \textbf{83.2} & \textbf{84.1} \\
\end{tabular}
\vspace{.3em}
\caption{\textbf{End-to-end fine-tuning} accuracy (\%) in ImageNet-1k.
\label{tab:e2e}
}
\vspace{-1.5em}
\end{table}

\paragraph{End-to-end fine-tuning.} Table~\ref{tab:e2e} reports end-to-end fine-tuning results. We use the DeiT codebase \cite{Touvron2020} and all its default settings unless specified. MoCo~v3 achieves \textbf{83.2\%} with ViT-B under 150-epoch fine-tuning, substantially better than DeiT's 81.8\% at 300 epochs.

In addition, MoCo~v3 has \textbf{84.1\%} with ViT-L when fine-tuned for only 100 epochs with a drop path rate of 0.5. This short schedule demonstrates the effectiveness of MoCo pre-training. We have also found DeiT-L diverges under its default settings, and a different solution may be needed.

\subsection{Transfer Learning}

In Table~\ref{tab:transfer} we evaluate transfer learning. We study the four downstream datasets as in \cite{Dosovitskiy2021}.
We fine-tune the models end-to-end, also following \cite{Dosovitskiy2021}.

Our self-supervised ViT has better transfer learning accuracy when the model size increases from ViT-B to \mbox{ViT-L}, yet it gets saturated when increased to \mbox{ViT-H}. As a comparison, the ImageNet-supervised ViT in \cite{Dosovitskiy2021} becomes saturated or overfitted starting at \mbox{ViT-L}. Our self-supervised ViT achieves \emph{better} results than its ImageNet-supervised counterparts in three of these four datasets.

The overfitting is more prominent when training the big ViT models from scratch in these small datasets: the accuracy in general decreases with bigger ViT. We also find that the from-scratch ViT results are much worse than their ResNet-counterparts (\cf, Table 8 in \cite{Chen2020}) in these small datasets. 
This suggests that if data are not enough, it is \emph{difficult} for ViT to learn representations in the lack of inductive biases. Self-supervised pre-training can close this gap and largely reduce overfitting in small datasets.

Finally, we note that with supervised pre-training in \mbox{bigger} datasets (ImageNet-21k or JFT-300M), the ViT results in \cite{Dosovitskiy2021} can be better than ours when transferring to these small datasets. A potential future work is to perform self-supervised pre-training for big ViT models in bigger data. This is analogous to the trajectory of unsupervised pre-training in NLP in the past years \cite{Radford2018,Devlin2019,Radford2019,Brown2020}, \ie, both models and datasets are scaled up. 

\section{Conclusion}

We have explored training ViT in the recently popular self-supervised frameworks. Our comparisons concern several aspects, including ViT \vs convolutional networks, supervised \vs self-supervised, and contrastive learning \vs masked auto-encoding. We report positive evidence as well as challenges, open questions, and opportunities. We hope our empirical study will be useful for the community to close the gap of pre-training between vision and language.



\paragraph{Postscript.}
After the first version of this manuscript, an author of BiT \cite{Kolesnikov2020}
and ViT \cite{Dosovitskiy2021}, \mbox{Lucas} Beyer, echoed that ``\emph{the exact same behaviour}'' was observed for supervised \mbox{BiT-ResNet} in ImageNet-21k. The instability problem can be more general than the scope of this paper.

\appendix

\section{Additional Implementation Details}

\paragraph{Data augmentation.} We follow the good practice in existing works \cite{Wu2018a,He2020,Chen2020,Grill2020}. Our augmentation policy includes random resized cropping, horizontal flipping, color jittering \cite{Wu2018a}, grayscale conversion \cite{Wu2018a}, blurring \cite{Chen2020}, and solarization \cite{Grill2020}. We take two 224$\times$224 crops for each image in each iteration.

\paragraph{BatchNorm.} We use SyncBN as our default BatchNorm implementation, following \cite{Chen2020}.
When BN is used, there are two options on batching: \textbf{(i)} all samples \emph{and} crops are in the same batch, \ie, BN is over 4096$\times$2 crops for 4096 images; and \textbf{(ii)} only different images are in the same batch, \ie, the two crops of the same image are separately forwarded in two 4096 batches. We notice that the code of SimCLR \cite{Chen2020} adopts the former option, while the code in BYOL \cite{Grill2020} adopts the latter. The pseudo-code in our Alg.~\ref{alg:code} implies that we adopt the latter.
The BN batching size influences the gradient variance, and the two implementations should lead to different results.

\paragraph{AdamW implementation.}
We notice that in PyTorch and JAX, the weight decay in AdamW is implemented as \mbox{``$\text{-}lr * wd * \text{weight}$'}' (consistent with \cite{Loshchilov2019}), but in \mbox{TensorFlow} it is implemented as ``$\text{-}wd * \text{weight}$'', and $wd$ needs to be scaled beforehand.\footnotemark~In our TPU/TensorFlow code, we follow the version consistent with \cite{Loshchilov2019}.

\footnotetext{{\fontsize{5.2pt}{1em}\selectfont\url{https://www.tensorflow.org/addons/api_docs/python/tfa/optimizers/AdamW}}}

\paragraph{MLP heads in BYOL and SwAV.} In our BYOL+ViT implementation, the projection/prediction MLP heads have BN in their hidden layers, but not in their output layers, which faithfully follow \cite{Grill2020}. In our SwAV+ViT implementation, we use no BN in the MLP heads, which is a configuration that performs the best in our SwAV experiments.

\paragraph{kNN monitor.} The kNN monitor \cite{Wu2018a} is a widely used tool in self-supervised learning research. The kNN evaluation was often performed sparsely, \eg, once per epoch. We notice that this may hide the sudden ``dips'' in the curves.

To better reveal the sudden changes, we monitor the kNN performance more densely, \eg, every tens of iterations. This is prohibitive even though the kNN classifier does not need training. We adopt a few approximations to make it feasible. We maintain a small memory bank \cite{Wu2018a} (whose length is 10\% of ImageNet) for the purpose of kNN search. This memory bank is updated per iteration by the features from the training samples (which are augmented images). This memory bank is maintained as a queue similar to \cite{He2020}, so it requires no extra feature extraction. We use the features from the class token for kNN monitoring, so the monitor is independent of the choice of the head.
Other details follow the kNN implementation in \cite{Wu2018a}. We find that this approximate kNN monitor is sufficient to reflect the stability of training.

\paragraph{MoCo~v3 for ResNet-50.} The implementation follows the good practice in recent works \cite{Chen2020,Grill2020}. \mbox{It uses the LARS} optimizer \cite{You2017} with a 4096 batch \cite{Chen2020}, $lr{=}0.3$, $wd{=}1.5e\text{-}6$. The temperature is $\tau{=}1.0$. The encoder $f_k$'s momentum coefficient is $m{=}0.996$ and increases to 1 with a cosine schedule \cite{Grill2020}. The augmentation is the same as described above.

{
\fontsize{8.5pt}{1em}\selectfont
\bibliographystyle{ieee_fullname}
\bibliography{moco_vit.bib}
}

\end{document}